\documentclass[11pt,x11names]{article}
\usepackage{latexsym,amssymb,amsmath,amsthm,amsfonts}
\usepackage[dvipsnames]{xcolor}
\usepackage{graphicx,shadow}
\usepackage{float}
\usepackage[numbers]{natbib}

\usepackage[a4paper,
margin=1in]{geometry}
\usepackage{setspace}
\usepackage{mathrsfs}
\usepackage{caption,subcaption}
\usepackage{soul}
\usepackage{colortbl} 
\usepackage{multirow} 
\usepackage{epstopdf}
\usepackage[hidelinks]{hyperref}
\usepackage[boxed,ruled,linesnumbered]{algorithm2e}
\usepackage{enumitem}
\usepackage[noabbrev,capitalise,nameinlink]{cleveref}
\hypersetup{colorlinks={true},linkcolor={MidnightBlue},citecolor=ForestGreen}

\usepackage[dvipsnames]{xcolor}
\usepackage[boxed,ruled,linesnumbered]{algorithm2e}
\usepackage{cleveref}

\numberwithin{equation}{section}

\def\argmin{\displaystyle\mathop {\mbox{argmin}}}

\newtheorem{assumption}{Assumption}

\newtheorem{theorem}{Theorem}[section]
\newtheorem{lemma}{Lemma}[section]
\newtheorem{corollary}{Corollary}[section]

\theoremstyle{definition}
\newtheorem{example}{Example}[section]
\newtheorem{definition}{Definition}[section]
\newtheorem{remark}{Remark}[section]

\def\prox{\displaystyle\mathop {\mbox{\rm prox}}}      

\def\dom{\displaystyle\mathop {\mbox{\rm dom}}}    

\def\dist{\displaystyle\mathop {\mbox{\rm dist}}}  

\def\sup{\displaystyle\mathop {\mbox{\rm sup}}}

\def\argmin{\displaystyle\mathop {\mbox{\rm argmin}}}

\def\real{\mathbb R}

\newcommand{\by}{{\bf y}}

\newcommand{\br}{{\bf r}}

\newcommand{\bA}{{\bf A}}

\newcommand{\C}{{\mathcal{C}}}

\newcommand{\T}{{\mathcal{T}}}

\DeclareMathOperator{\reg}{Reg}		
\DeclareMathOperator{\Reg}{Reg}		
\DeclareMathOperator{\DReg}{DReg}		

\newcommand{\keywords}[1]{%
	\vspace{0.5cm}
	\noindent\textbf{Keywords:} #1
}

\title{A Stationarity-Based Regret Framework for Time-Varying Nonconvex Composite Optimization}

\author{Nadav Hallak\thanks{Data and Decision Sciences, Technion, Haifa, Israel. \url{ndvhllk@technion.ac.il}, ORCID0000-0002-0045-6636}}
\date{}

\begin{document}
\maketitle

\begin{abstract}
This paper studies time-varying nonconvex composite optimization, where decisions are made sequentially and classical value-based regret notions are generally ill-posed.
We introduce a stationarity-based performance framework defined through local regret measures
built from first-order optimality conditions, and identify minimal structural requirements
under which such measures constitute sound performance criteria.
This abstraction encompasses canonical optimality measures, including the proximal-gradient residual, and collapses to classical dynamic regret in convex settings.
We then focus on the proximal-gradient residual as a concrete and algorithmically meaningful instantiation for general composite problems.
In convex problems, we show that the resulting local regret is equivalent to standard dynamic regret.
In nonconvex settings, we establish that sublinear stationarity-based regret is unattainable without temporal regularity, and introduce a gradient-variation budget that captures this necessity.
Under this model, we analyze online proximal-gradient methods and prove sublinear regret guarantees in both nested multi-call and fully incremental single-call oracle regimes, highlighting the tradeoff between per-round computational effort and achievable performance.
Finally, we demonstrate how the proposed framework extends to  nonconvex multiplayer
games by interpreting equilibrium-seeking dynamics through stationarity-based regret, and
corroborate the theoretical findings with numerical experiments.
\end{abstract}

\keywords{
	composite optimization,
	nonconvex,
	online,
	dynamic regret,
	proximal-gradient
}


\section{Introduction}
\subsection{The Time-Varying Nonconvex Composite Optimization Model}
\label{composite model}

We study a time-varying (online) composite optimization problem over a finite horizon
$t=1,\ldots,T$, arising in both online decision-making and
{online adversarial optimization}.
At each time $t\in [T]$, the decision maker selects a vector $x_t$ from a fixed feasible
region.
A (possibly nonconvex) smooth loss function $f_t$ is then revealed, and performance
is evaluated with respect to the instantaneous composite objective
\begin{equation}
	\label{prob:1}
	\tag{P}
	\min_{x\in\real^n} \left\{ F_t(x) := f_t(x) + g(x) \right\}.
\end{equation}
Here, $g$ is a time-invariant (possibly nonsmooth) regularization term encoding
structural constraints and desired solution properties, while the sequence
$\{f_t\}$ captures temporal variation in the objective.
No probabilistic assumptions are imposed on the evolution of $\{f_t\}$, allowing
for adversarial but regularity-constrained dynamics.

\medskip
Throughout, we impose the following standing assumptions.

\begin{itemize}
	\item The function $g:\real^n \to \real\cup\{\infty\}$ is proper, closed, and convex, and admits
	a decomposition
	\[
	g = g_0 + \delta_{\mathcal C},
	\]
	where $\mathcal C\subset\real^n$ is a nonempty compact convex set and
	$g_0:\real^n\to\real$ is convex and Lipschitz continuous on $\mathcal C$.
	We denote by $D$ the diameter of $\mathcal C$.

	\item For each $t\in[T]$, the loss function $f_t:\real^n\to\real$ is continuously differentiable
	on $\mathcal C$ and $L_f$-smooth, i.e.,
	\[
	\|\nabla f_t(x)-\nabla f_t(y)\|\le L_f\|x-y\|,
	\quad \forall x,y\in\mathcal C.
	\]

	\item There exists a constant $M>0$ such that
	\[
	|f_t(x)|,\ |F_t(x)|,\ \|\nabla f_t(x)\|\le M,
	\quad \forall x\in\mathcal C,\ \forall t\in[T].
	\]
\end{itemize}

The boundedness conditions above are standard in online optimization and
follow in part from the compactness of the feasible set.

\medskip

Many sequential decision problems in operations research, machine learning, and data science naturally give rise to online optimization models, in which decisions are made sequentially against time-varying and potentially nonconvex objectives.
Such nonstationarity may stem from evolving data within a fixed decision framework or from structural changes driven by exogenous factors.
Models of this form are ubiquitous in dynamic optimization and decision-making; see, for example, \cite{BMP90,KY03,CL06,H16,BS13}.

These settings share two features that are central to this work:
(i) the objective functions evolve over time, and
(ii) value-based comparisons to global minimizers are generally ill posed.
As a result, meaningful performance assessment requires alternative criteria that remain both tractable and informative under nonstationarity and nonconvexity.

\medskip
With this modeling framework in place, we briefly recall the classical
\emph{dynamic regret} benchmark from online convex optimization, which will
serve as a reference point for assessing the soundness of the stationarity-based
measures introduced later in \Cref{sec:The Local Regret Measure}.
Informally, the key idea is to replace comparisons to per-round minimizers---which
may be ill defined in nonconvex problems---by per-round optimality measures that
quantify first-order suboptimality.

Concretely, let $\{x_t\}_{t=1}^T \subseteq \mathcal C$ denote the sequence of
decisions generated by an online algorithm, and let $\{\mathcal M_t\}$ be a
 sequence of tractable optimality measures (e.g., the gradient-mapping \cite{B14}).
The associated \emph{local regret} of $\{x_t\}_{t=1}^T$ is defined as
\[
\Reg_{\mathcal M}(T)
\;:=\;
\sum_{t=1}^T \mathcal M_t(x_t),
\]
and provides a cumulative measure of deviation from first-order optimality over
time.

\paragraph{Dynamic benchmark and regret.}
For each $t\in[T]$, let
\begin{equation}\label{eq:xtstar-def}
x_t^\star \in \arg\min_{x\in\mathbb{R}^n} F_t(x)
\end{equation}
denote a (possibly nonunique) per-round minimizer.
The dynamic regret with respect to the comparator sequence
$\{x_t^\star\}_{t=1}^T$ is given by
\begin{equation}\label{eq:dyn-regret-def}
\DReg(T)
\;:=\;
\sum_{t=1}^T \bigl(F_t(x_t)-F_t(x_t^\star)\bigr).
\end{equation}
When the objective is time-independent, i.e., $F_t \equiv F$ for all $t$, this
definition reduces to the standard cumulative suboptimality gap relative to a
minimizer of $F$.

In nonconvex settings, however, global minimizers may be unattainable, rendering dynamic regret unsuitable as a primary
performance benchmark and motivating the use of stationarity-based alternatives.


\subsection{Contribution and Main Results}

This paper studies time-varying nonconvex composite optimization through
stationarity-based regret, a performance measure defined via violations of
first-order optimality conditions.
We establish fundamental limits for such measures, characterize the role of
temporal regularity, and analyze the algorithmic consequences under different
oracle-access models.

Our main contributions are as follows.
\begin{itemize}

\item \textbf{Stationarity-based regret.}
We introduce a local regret formulation for time-varying (online) nonconvex composite optimization based
on first-order optimality measures and identify minimal conditions under which
it constitutes a sound performance criterion.
In convex settings, we show that this regret upper bounds classical dynamic
regret.

\item \textbf{Impossibility without temporal regularity.}
For the proximal-gradient optimality measure, we prove that sublinear
stationarity-based regret is unattainable in general unless the loss sequence
satisfies explicit temporal regularity conditions.

\item \textbf{Online proximal-gradient guarantees.}
Under a gradient-variation budget, we analyze online proximal-gradient methods
and establish sublinear stationarity-based regret bounds under both nested
multi-call and fully incremental single-call oracle models, together with
explicit oracle-complexity guarantees.

\item \textbf{Structured extensions.}
We show that the framework extends to nonconvex multiplayer games,
where stationarity-based regret quantifies approximate first-order Nash
equilibria over time, and illustrate the theory numerically.

\end{itemize}

To the best of our knowledge, this work provides the first unified and tractable framework
for time-varying (online) nonconvex composite optimization that yields stationarity-consistent performance guarantees and recovers classical dynamic regret in convex settings.

\subsection{Related Work}
\label{sec:Related Work}

Online optimization is a well-established area spanning optimization, game theory, and
operations research.
The classical theory focuses primarily on convex settings, where regret notions based on
function values are both meaningful and tractable; see, for example,
\cite{H16,CO21,HSLZ21}.

\paragraph{Dynamic regret in convex online optimization.}
In online convex optimization, a central benchmark for nonstationary environments is
\emph{dynamic regret}, which evaluates an algorithm by comparison to the sequence of
per-round minimizers.
This notion was introduced by \cite{Zinkevich03} and has since been studied extensively
under assumptions that restrict the temporal variation.
Such analyses typically impose regularity through path-length, function-variation, or
gradient-variation conditions, under which sublinear dynamic regret guarantees can be
established for smooth convex losses; see, for example,
\cite{Zinkevich03,BGZ15,JRSS15} and the references therein.
These results provide a well-established foundation for performance evaluation in convex, time-varying optimization.

\paragraph{Value-based regret in nonconvex settings.}
Extending dynamic regret guarantees beyond convexity is substantially more challenging,
as global minimizers are generally intractable even in static nonconvex problems.
One line of work nevertheless pursues \emph{value-based} regret in nonconvex online learning
by relying on oracle-based methods or additional structural assumptions.
For example, \cite{XZ24} study dynamic regret for nonconvex losses under a functional
variation budget and propose restarted variants of follow-the-perturbed-leader, obtaining
sublinear dynamic regret guarantees via optimization-oracle primitives.
Related approaches mitigate nonconvexity by assuming access to exact or approximate offline
solvers \cite{AGH19,SN20} or by imposing conditions such as the Polyak--\L{}ojasiewicz
inequality \cite{KMDD22}, which effectively restore global tractability.
While these methods enable value-based regret guarantees, they rely on assumptions that
significantly restrict the class of admissible problems and depart from standard
first-order methodology in nonconvex  optimization models.

\paragraph{Stationarity-based local regret.}
A different line of work replaces global optimality benchmarks with \emph{local performance
measures} based on first-order stationarity conditions.
Pioneered by \cite{HSZ17} and further developed in
\cite{HMC21,GZL23,TNHSB24,LSJLS24}, this approach defines local regret using the norm, often
squared, of a stationarity measure evaluated on a sliding-window average of past losses.
These notions are compatible with first-order oracle models and recover stationarity
guarantees familiar from offline nonconvex optimization.

Despite their algorithmic appeal, sliding-window local regret notions lack a principled
connection to classical online benchmarks.
In particular, it has remained unclear whether sublinear sliding-window local regret
implies sublinear dynamic regret, even in convex problems.
Moreover, the reliance on growing time windows introduces explicit horizon dependence and
requires repeated access to past gradients, which may be incompatible with standard online
oracle models.

\paragraph{Positioning of the present work}
We show that this disconnect is structural rather than technical: sublinear sliding-window
local regret can coexist with linear dynamic regret, even under convexity.
Motivated by this observation, we introduce a gradient-mapping--based notion of local regret
defined directly with respect to the instantaneous online losses.

\subsection{Mathematical Preliminaries and Notation}
\label{sec:prelim}

The proximal-gradient (prox-grad) mapping and its associated optimality measure, namely the \emph{gradient mapping} (see
\cite[Section~10.3.2]{B17}), are defined as follows.
For a search direction $d\in\real^n$ and step size $\eta>0$, define the proximal mapping
associated with $g$ by
\begin{equation}\label{eq:prox-res}
	\T_{\eta}^{g}(x;d)
	\;:=\;
	\prox_{\eta g}\!\left(x-\eta d\right)
	\;=\;
	\argmin_{z\in\real^n}
	\left\{
	\eta g(z) + \tfrac{1}{2}\|x-\eta d-z\|^2
	\right\}.
\end{equation}
The corresponding proximal-gradient residual is defined as
\begin{equation}\label{eq:pg-residual}
	\mathcal{P}_{\eta}^{g}(x;d)
	\;:=\;
	\frac{1}{\eta}\bigl(x-\T_{\eta}^{g}(x;d)\bigr).
\end{equation}
When $d=\nabla f(x)$, the residual
$\mathcal{P}_{\eta}^{g}(x;\nabla f(x))$ coincides with the standard gradient mapping
used to quantify first-order stationarity in composite (possibly nonconvex) optimization:
$\mathcal{P}_{\eta}^{g}(x;\nabla f(x))=0$ if and only if
$0\in\nabla f(x)+\partial g(x)$; see \cite[Chapter~10]{B17}.

\medskip
\noindent
Throughout the paper, we rely on standard smoothness properties and basic facts about
proximal operators.
For completeness, we state here the specific results that are used  in the
sequel.

\begin{lemma}[Descent lemma]\label{lemma:des}
Let $f:\real^n\to\real$ be continuously differentiable with $L_f$-Lipschitz continuous
gradient on a convex set $C\subseteq\real^n$. Then for any $x,y\in \C$,
\[
f(y)
\le
f(x)+\langle \nabla f(x),y-x\rangle+\frac{L_f}{2}\|y-x\|^2.
\]
\end{lemma}

\begin{lemma}[Sufficient decrease of the proximal-gradient step]
\label{lemma:sufficient decrease}
Let $f:\real^n\to\real$ be continuously differentiable with $L_f$-Lipschitz continuous
gradient and let $g:\real^n\to(-\infty,+\infty]$ be proper, closed, and convex.
Fix $\eta\in(0,1/L_f]$ and define
\[
x^{+}:=\prox_{\eta g}\bigl(x-\eta\nabla f(x)\bigr).
\]
Then
\[
f(x^{+})+g(x^{+})
\le
f(x)+g(x)
-
\frac{1-\eta L_f}{2\eta}\|x^{+}-x\|^2.
\]
\end{lemma}

\begin{theorem}[Nonexpansivity of the proximal operator {\cite[Theorem~6.42]{B17}}]
\label{thm:nonexpansivity}
Let $g:\real^n\to(-\infty,+\infty]$ be proper, closed, and convex. Then for any
$x,y\in\real^n$ and any $\eta>0$,
\[
\bigl\|\prox_{\eta g}(x)-\prox_{\eta g}(y)\bigr\|
\le
\|x-y\|.
\]
\end{theorem}

\medskip
\noindent
The following lemma summarizes stability properties of the proximal-gradient mapping with
respect to perturbations in the direction argument; it plays a key role in controlling the
effect of temporal variation in the online setting.

\begin{lemma}[Stability of the proximal-gradient mapping]
\label{lem:T-eta-sensitivity}
Let $x\in\real^n$, $\eta>0$, and $d_1,d_2\in\real^n$. 
Then
\[
\bigl\|\mathcal{P}^g_{\eta}(x;d_1+d_2)\bigr\|
\le
\bigl\|\mathcal{P}^g_{\eta}(x;d_1)\bigr\|
+
\|d_2\|.
\]
\end{lemma}

\begin{proof}
Set $z_1:=\T^g_\eta(x;d_1)$ and $z_{12}:=\T^g_\eta(x;d_1+d_2)$. Then
\[
\mathcal{P}^g_\eta(x;d_1+d_2)-\mathcal{P}^g_\eta(x;d_1)
=
\frac{1}{\eta}(z_1-z_{12}).
\]
By \Cref{thm:nonexpansivity},
\[
\|z_1-z_{12}\|
\le
\|\eta d_2\|
=
\eta\|d_2\|,
\]
and the claim follows from the triangle inequality.
\end{proof}

\section{The Local Regret Measure}
\label{sec:The Local Regret Measure}

This section introduces the performance benchmark used throughout the paper.
Our objective is to assess how well an online algorithm tracks first-order
stationary behavior of a time-varying composite objective.
To this end, we use a notion of \emph{local regret} as a cumulative measure of
stationarity drift over time.

Two results motivate this choice.
In convex settings, \Cref{thm:regret-local-to-dynamic} shows that local regret is
equivalent to classical dynamic regret up to constant factors.
More generally, \Cref{lem:necessity_variation_general} shows that without
temporal regularity, sublinear control of stationarity drift is not achievable.

\subsection{Local regret via optimality measures}
\label{subsec:local-regret-opt-measures}

We define local regret through violations of first-order optimality conditions associated
with the time-varying composite problems
\[
\min_x \; F_t(x) := f_t(x)+g(x).
\]
Specifically, we quantify performance via suitable optimality measures that capture
deviation from stationarity at each time step.

To formalize this notion, consider a sequence of nonnegative mappings
\[
\{\mathcal M_t:\dom g \to \mathbb R_+\}_{t\ge1},
\]
where $\mathcal M_t(x)$ measures the violation of first-order stationarity for the composite
objective $F_t$.
In the stationary case, i.e., $f_t \equiv f$, such mappings coincide with standard optimality
measures used in nonconvex optimization; see, e.g., \cite{B17}.

\begin{definition}[Admissible online optimality measure]
\label{def:admissible-measure}
We say that $\mathcal M_t$ is an \emph{admissible online optimality measure} if it satisfies
the following two properties.
\begin{enumerate}[label=(P\arabic*)]

\item \textbf{Stationarity.}
$\mathcal M_t(x)=0$ if and only if $x$ is a first-order stationary point of $F_t$.

\item \textbf{Compatibility with convex optimization.}
If $F_t$ is convex, then there exists a constant $C>0$ such that for all $x\in\dom g$,
\[
F_t(x)-F_t(x_t^\star)
\;\le\;
C\,\mathcal M_t(x),
\qquad
x_t^\star\in\arg\min F_t .
\]

\end{enumerate}
\end{definition}

Given an admissible optimality measure $\{\mathcal M_t\}$, the associated
\emph{local regret} of a decision sequence $\{x_t\}_{t=1}^T$ is defined as
\begin{equation}\label{eq:Reg-general}
\Reg_{\mathcal M}(T)
\;:=\;
\sum_{t=1}^T \mathcal M_t(x_t).
\end{equation}
The following result shows that property \textup{(P2)} is sufficient to recover classical
dynamic regret guarantees in convex settings.
As a consequence, the local regret defined in \eqref{eq:Reg-general} constitutes a sound
performance measure: it vanishes exactly at first-order stationary points and, when the
losses are convex, it upper bounds dynamic regret up to constant factors.

\begin{lemma}[Dynamic regret control under convex compatibility]
\label{lem:conv-compat-implies-dreg}
Suppose that $F_t$ is convex for all $t\in[T]$ and that the optimality measure
$\{\mathcal M_t\}_{t=1}^T$ satisfies property \textup{(P2)} with constant $C>0$.
Then for any decision sequence $\{x_t\}_{t=1}^T\subseteq\dom g$,
\[
\DReg(T)
=
\sum_{t=1}^T\bigl(F_t(x_t)-F_t(x_t^\star)\bigr)
\;\le\;
C\,\Reg_{\mathcal M}(T).
\]
In particular, sublinear local regret implies sublinear dynamic regret.
\end{lemma}

\begin{proof}
Apply \textup{(P2)} with $x=x_t$ and sum over $t=1,\ldots,T$.
\end{proof}

\medskip
\noindent
The admissible-measure abstraction serves to isolate minimal structural requirements for a
sound stationarity-based regret notion.
We next present canonical instances of admissible online optimality measures.

\begin{lemma}[Proximal-gradient admissibility]
\label{thm:pg-admissible}
Under the standing assumptions, the proximal-gradient mapping
\[
\mathcal M_t^{\mathrm{PG}}(x)
=
\|\mathcal P^g_\eta(x;\nabla f_t(x))\|
\]
defines an admissible online optimality measure in the sense of
\Cref{def:admissible-measure}.
\end{lemma}

\begin{proof}
Property~\textup{(P1)} follows from the characterization of first-order stationarity via
the proximal-gradient mapping; see \Cref{lem:gm-fermat}.
Property~\textup{(P2)} is established in
\Cref{thm:regret-local-to-dynamic}.
\end{proof}

\begin{corollary}[Projected-gradient admissibility]
When $g\equiv\delta_{\mathcal C}$, the proximal-gradient
mapping reduces to the projected-gradient mapping, and the online optimality measure to
\[
\mathcal M_t^{\mathrm{Proj}}(x)
\;:=\;
\bigl\|x-\Pi_{\mathcal C}\bigl(x-\eta\nabla f_t(x)\bigr)\bigr\|.
\]
Admissibility follows directly as a special case of
\Cref{thm:pg-admissible}.
\end{corollary}

\begin{lemma}[Frank--Wolfe admissibility]
\label{thm:fw-admissible}
Assume $g\equiv\delta_{\mathcal C}$. 
Then the Frank--Wolfe gap
\[
\mathcal M_t^{\mathrm{FW}}(x)
\;:=\;
\max_{z\in\mathcal C}
\langle \nabla f_t(x),\, x-z\rangle,
\qquad x\in\mathcal C.
\]
defines an admissible online
optimality measure.
\end{lemma}

\begin{proof}
Property~\textup{(P1)} follows from the standard first-order optimality condition for
constrained smooth optimization.
If $f_t$ is convex, then for any $x\in\mathcal C$ and any minimizer
$x_t^\star\in\arg\min_{\mathcal C} f_t$, convexity implies
\[
f_t(x)-f_t(x_t^\star)
\le
\langle \nabla f_t(x),\,x-x_t^\star\rangle
\le
\max_{z\in\mathcal C}\langle \nabla f_t(x),\,x-z\rangle
=
\mathcal M_t^{\mathrm{FW}}(x),
\]
which establishes property~\textup{(P2)}.
\end{proof}

In a similar manner as for the FW-gap, we can show that the optimality measure associated with classical Feasible-Direction methods is also admissible.

\begin{lemma}[Feasible-direction admissibility]
\label{lem:fd-admissible}
Assume $g\equiv\delta_{\mathcal C}$.
Define the feasible-direction optimality measure
\[
\mathcal M_t^{\mathrm{FD}}(x)
\;:=\;
\max_{\substack{z\in\mathcal C\\ z\neq x}}
\frac{-\langle \nabla f_t(x),\, z-x\rangle}{\|z-x\|},
\qquad x\in\mathcal C.
\]
Then $\{\mathcal M_t^{\mathrm{FD}}\}_{t=1}^T$ defines an admissible online
optimality measure.
\end{lemma}

\medskip
\noindent
Among these instances, the proximal-gradient mapping is the only one that naturally extends
to general composite models and remains directly connected to the first-order optimality
conditions,  known as the Fermat conditions, of nonconvex problems (cf.~\Cref{lem:gm-fermat}).
For this reason, it constitutes the primary focus of the subsequent analysis.

\medskip

We conclude this subsection by contrasting \eqref{eq:Reg-general} with \emph{sliding-window} local regret notions
\cite{HSZ17,HMC21}, which evaluate stationarity with respect to the window-averaged loss
$S_{w,t} := \tfrac{1}{w}\sum_{i=t-w+1}^{t} f_i$.
Although such criteria have gained  popularity
\citep{GZL23,TNHSB24,LSJLS24} and are often regarded as the standard benchmark in nonconvex online optimization, they are inherently designed to assess the local stability of smoothed objectives and, in general, do not align with dynamic-regret benchmarks.

\begin{lemma}[Sliding-window local regret does not control dynamic regret]
\label{lem:failure-window}
There exist a convex online optimization problem and a sequence of loss functions
$\{f_t\}_{t=1}^T$ such that:
\begin{enumerate}
\item the sliding-window local regret (with window size $w$) is sublinear in $T$;
\item the dynamic regret with respect to the per-round minimizers
$\{x_t^\star\}$ grows linearly in $T$.
\end{enumerate}
\end{lemma}

\begin{proof}
We construct a simple convex instance in which stationarity of the window-averaged loss
is maintained, while the per-round minimizers oscillate, inducing linear dynamic regret.

Let $\C=[-1,1]$, let $g=\delta_\C$, and fix an even window size $w$.
Define the loss sequence $\{f_t\}_{t=1}^T$ by
\[
f_t(x) := (-1)^t M x, \qquad t=1,2,\ldots,T,
\]
for some constant $M>0$.
Each $f_t$ is linear and hence convex.

Consider the sliding-window averaged loss
\[
S_{w,t}(x) := \frac{1}{w}\sum_{i=t-w+1}^t f_i(x).
\]
Since $w$ is even and the sequence $\{f_t\}$ alternates in sign, each window of length $w$
contains exactly $w/2$ terms equal to $Mx$ and $w/2$ terms equal to $-Mx$, and therefore
\[
S_{w,t}(x) \equiv 0 \qquad \text{for all } t \ge w.
\]
Consequently, every point $x\in[-1,1]$ is a stationary point of $S_{w,t}$ for all $t\ge w$.
In particular, the constant decision sequence $x_t \equiv 0$ yields zero sliding-window
stationarity residual for all $t\ge w$, and hence the sliding-window local regret is $O(w)$,
and thus sublinear in $T$.

On the other hand, the per-round minimizer
\[
x_t^\star \in \arg\min_{x\in[-1,1]} f_t(x)
\]
satisfies $f_t(x_t^\star) = -M$ for all $t$, whereas $f_t(0)=0$.
Therefore, the dynamic regret with respect to $\{x_t^\star\}$ satisfies
\[
\sum_{t=1}^T \bigl(f_t(0) - f_t(x_t^\star)\bigr)
= MT,
\]
which grows linearly in $T$.
This shows that sliding-window local regret does not control dynamic regret, even for convex losses.
\end{proof}

\medskip
\Cref{lem:failure-window} shows that sliding-window local regret may remain small even when dynamic regret grows linearly, even for convex losses.
This demonstrates that property (P2) is not merely technical and justifies the formulation in \eqref{eq:Reg-general}, which evaluates stationarity directly with respect to the instantaneous online losses rather than their temporal averages.
In particular, sliding-window regret is ruled out as a sound performance benchmark, even in convex settings.
The remaining issue is therefore not the choice of a regret notion, but the conditions under which any sound notion can be controlled.

\subsection{Local regret via the proximal-gradient mapping}
\label{subsec:local-regret-prox}

We now specialize the abstract framework of
\Cref{subsec:local-regret-opt-measures}
to the proximal-gradient mapping, which provides a natural optimality measure for
composite objectives.

Throughout this section, the stepsize $\eta$ is treated as a fixed problem-level parameter.
First-order stationarity of $F_t$ is characterized by the Fermat condition
$0\in \nabla f_t(x)+\partial g(x)$, and the proximal-gradient mapping yields a
computable residual quantifying violations of this condition.

\begin{definition}[Proximal-gradient-mapping--based local regret]
\label{def:local regret}
Given \eqref{prob:1}, the \emph{local regret with respect to a stepsize
$\eta \in (0,1/L_f]$} of a decision sequence
$\{x_t\}_{t=1}^T\subseteq\dom g$ is defined as
\begin{equation}\label{eq:local regret}
\reg(T)
:=
\sum_{t=1}^{T}
\bigl\|
\mathcal{P}^g_{\eta}\bigl(x_t;\nabla f_t(x_t)\bigr)
\bigr\|.
\end{equation}
\end{definition}

The non-squared aggregation is essential for convex compatibility and avoids degeneracies
that arise in squared or sliding-window stationarity criteria.

To interpret the gradient-mapping residual as a genuine measure of first-order stationarity,
we relate it to the classical Fermat residual associated with the composite objective.

\begin{lemma}[Gradient mapping bounds the Fermat residual]
\label{lem:gm-fermat}
Let $f:\real^n\to\real$ have $L_f$-Lipschitz continuous gradient and let
$g:\real^n\to(-\infty,+\infty]$ be proper, closed, and convex.
Fix $\eta>0$ and define
$x^{+} := \prox_{\eta g}\bigl(x-\eta\nabla f(x)\bigr)$.
Then
\[
\dist\bigl(0,\nabla f(x^{+})+\partial g(x^{+})\bigr)
\;\le\;
(1+\eta L_f)\,
\bigl\|\mathcal{P}^g_{\eta}(x;\nabla f(x))\bigr\|.
\]
\end{lemma}

\begin{proof}
By optimality of the proximal step,
$\frac{1}{\eta}(x-x^+)-\nabla f(x)\in\partial g(x^+)$.
Thus
\[
\dist\bigl(0,\nabla f(x^+)+\partial g(x^+)\bigr)
\le
\|\nabla f(x^+)-\nabla f(x)\|
+
\frac{1}{\eta}\|x-x^+\|.
\]
Using $L_f$-smoothness and
$\|x-x^+\|=\eta\|\mathcal{P}^g_\eta(x;\nabla f(x))\|$
yields the claim.
\end{proof}

\begin{corollary}[Local regret implies cumulative Fermat violation]
\label{cor:local-to-fermat}
Let $\{x_t\}_{t=1}^T$ be any sequence and define
$x_t^{+} := \prox_{\eta g}\bigl(x_t-\eta\nabla f_t(x_t)\bigr)$.
If $\reg(T)=o(T)$, then
\[
\sum_{t=1}^T
\dist\bigl(0,\nabla f_t(x_t^{+})+\partial g(x_t^{+})\bigr)
=
o(T).
\]
In particular, the average first-order stationarity violation along the proximal points
$\{x_t^{+}\}$ vanishes.

\end{corollary}

We next position \eqref{eq:local regret} relative to classical benchmarks.
First, dynamic regret always controls a weighted sum of squared gradient-mapping
residuals, without any temporal regularity assumption (\Cref{lem:dyn-to-local}).
Second, in the convex case, local regret also controls dynamic regret
(\Cref{thm:regret-local-to-dynamic}), establishing equivalence up to constants.
Finally, some form of temporal regularity is necessary for sublinear local regret against
adaptive losses (\Cref{lem:necessity_variation_general}).

\begin{lemma}[Dynamic regret controls local regret]
\label{lem:dyn-to-local}
Assume that the standing conditions of the model hold.
Then, for any sequence $\{x_t\}_{t=1}^T\subseteq\dom g$,
\begin{equation}\label{eq:local-dyn-bound}
\sum_{t=1}^T \eta \,
\bigl\|
\mathcal{P}^g_{\eta}\bigl(x_t;\nabla f_t(x_t)\bigr)
\bigr\|^2
\;\le\;
2\,\DReg(T),
\end{equation}
where $\DReg(T)=\sum_{t=1}^T\bigl(F_t(x_t)-F_t(x_t^\star)\bigr)$.
\end{lemma}

\begin{proof}
Fix $t$ and set $x_t^{+}:=\prox_{\eta g}\bigl(x_t-\eta\nabla f_t(x_t)\bigr)$ and
$G_t:=\|\mathcal{P}^g_{\eta}(x_t;\nabla f_t(x_t))\| = \frac{1}{\eta}\|x_t-x_t^{+}\|$.
By sufficient decrease (\Cref{lemma:sufficient decrease}) and $\eta\le 1/L_f$,
\[
F_t(x_t)-F_t(x_t^{+})\;\ge\;\frac{\eta}{2} G_t^2.
\]
Since $x_t^\star$ minimizes $F_t$ over $\dom g$, $F_t(x_t^{+})\ge F_t(x_t^\star)$, hence
$F_t(x_t)-F_t(x_t^\star)\ge \frac{\eta}{2} G_t^2$.
Summing over $t$ yields \eqref{eq:local-dyn-bound}.
\end{proof}

The following corollary translates the cumulative relation in
\Cref{lem:dyn-to-local} into a pointwise stationarity guarantee, showing that
sublinear dynamic regret ensures the existence of a nearly stationary iterate.

\begin{corollary}[Best-iterate stationarity from dynamic regret]
\label{cor:dyn-to-stationarity}
Under the conditions of \Cref{lem:dyn-to-local}, let $G_t:=\|\mathcal{P}^g_{\eta}(x_t;\nabla f_t(x_t))\|$.
Then
\[
\min_{t\in[T]} G_t
\;\le\;
\left(\frac{2}{\eta T}\DReg(T)\right)^{1/2}.
\]
\end{corollary}

\medskip

\begin{remark}[From dynamic regret to cumulative stationarity]
\label{rem:dyn-to-local-to-regret}
\Cref{lem:dyn-to-local} bounds the weighted squared residual
$\sum_{t=1}^T \eta G_t^2$, where
$G_t := \|\mathcal{P}^g_{\eta}(x_t;\nabla f_t(x_t))\|$. This bound yields, by Cauchy--Schwarz,
\[
\reg(T)
= \sum_{t=1}^T G_t
\;\le\;
\sqrt{T}\Bigl(\sum_{t=1}^T G_t^2\Bigr)^{1/2}
\;\le\;
\sqrt{\frac{2T}{\eta}\,\DReg(T)}.
\]
\end{remark}

We conclude this subsection by establishing the converse implication in the convex case,
using a classical proximal-gradient inequality \cite[Theorem~10.16]{B17}.

\begin{corollary}
\label{cor:pg-ineq}
If $\eta\le 1/L_f$, then for any $x\in\dom g$ and any $x_t\in\dom g$,
\begin{equation}
\label{eq:1}
2\eta\Bigl(F_t(x)-F_t\!\left(\T_{\eta}^{g}\bigl(x_t;\nabla f_t(x_t)\bigr)\right)\Bigr)
\;\ge\;
\left\|x-\T_{\eta}^{g}\bigl(x_t;\nabla f_t(x_t)\bigr)\right\|^2
-\|x-x_t\|^2.
\end{equation}
\end{corollary}

\begin{proof}
	The proof follows technically from applying the proximal-gradient inequality {\cite[Theorem~10.16]{B17}}  to $f=f_t$ with $y=x_t$ and $L=1/\eta$.
\end{proof}

\begin{theorem}[Local regret controls dynamic regret in the convex case]
\label{thm:regret-local-to-dynamic}

Assume $f_t$ is convex.
Then for any sequence $\{x_t\}_{t=1}^T\subseteq\dom g$,
\begin{equation}\label{eq:local-to-dynamic}
\DReg(T)
\;\le\;
\left(D + \frac{M+L_g}{L_f}\right)
\Reg (T).
\end{equation}
Consequently, sublinear local regret implies sublinear dynamic regret.
\end{theorem}
\begin{proof}
Fix $t\in[T]$ and let $x_t^\star\in\arg\min F_t$.
Denote $x_t^+:=\T_{\eta}^g(x_t;\nabla f_t(x_t))$ and
$G_t:=\mathcal{P}^g_{\eta}(x_t;\nabla f_t(x_t))$ so that $x_t-x_t^+=\eta G_t$.

Apply \Cref{cor:pg-ineq} with $x=x_t^\star$ to obtain
\[
F_t(x_t)-F_t(x_t^\star)
\le
F_t(x_t)-F_t(x_t^+)
+\frac{1}{2\eta}\Bigl(\|x_t^\star-x_t\|^2-\|x_t^\star-x_t^+\|^2\Bigr).
\]

For the second term, using $a^2-b^2=(a+b)(a-b)$ and $\mathrm{diam}(\dom g)=D$ gives
\[
\|x_t^\star-x_t\|^2-\|x_t^\star-x_t^+\|^2
\le
(\|x_t^\star-x_t\|+\|x_t^\star-x_t^+\|)\|x_t-x_t^+\|
\le
2D\,\|x_t-x_t^+\|
=
2D\,\eta\|G_t\|,
\]
hence
\[
\frac{1}{2\eta}\Bigl(\|x_t^\star-x_t\|^2-\|x_t^\star-x_t^+\|^2\Bigr)
\le D\|G_t\|.
\]

For the first term, by $M$-Lipschitz continuity of $f_t$ and $L_g$-Lipschitz continuity of $g_0$ on $\dom g$,
\begin{align*}
	F_t(x_t)-F_t(x_t^+)
=
(f_t(x_t)-f_t(x_t^+))+(g(x_t)-g(x_t^+))
\le
(M+L_g)\|x_t-x_t^+\|
&=
(M+L_g)\eta\|G_t\|
\\ &\le
\frac{M+L_g}{L_f}\,\|G_t\|.
\end{align*}

Combining the bounds yields
\[
F_t(x_t)-F_t(x_t^\star)
\le
\left(D+\frac{M+L_g}{L_f}\right)\|G_t\|.
\]
Summing over $t=1,\ldots,T$ completes the proof.
\end{proof}

\begin{corollary}[Projected case]
If $g=\delta_\C$, then the same conclusion holds with the term
$L_g$ absent:
\[
\DReg(T)\le\left(D+\frac{M}{L_f}\right)\reg(T).
\]
\end{corollary}

The preceding results establish proximal-gradient local regret as a sound performance
measure: it characterizes first-order stationarity and, in convex settings, coincides
with classical dynamic regret up to constants.
Soundness alone, however, does not guarantee attainability.

In the absence of restrictions on the temporal evolution of the loss sequence, even this
stationarity-based criterion cannot be controlled.
The following lemma makes this precise by showing that an adaptive adversary can enforce
a constant proximal-gradient residual at every round, resulting in linear cumulative
local regret, even for benign composite models.

\begin{lemma}[Impossibility of sublinear proximal-gradient regret without temporal regularity]
\label{lem:necessity_variation_general}
Fix a stepsize $\eta>0$.
For any online algorithm that outputs $x_t\in\mathcal C$ before observing $f_t$,
there exists an adaptive sequence of linear loss functions with uniformly bounded
gradients such that the proximal-gradient mapping satisfies
\[
\bigl\|\mathcal{P}^g_{\eta}\bigl(x_t;\nabla f_t(x_t)\bigr)\bigr\|
\;\ge\; c
\qquad\text{for all }t\ge1,
\]
for some constant $c>0$ independent of $t$.
Consequently, the cumulative proximal-gradient local regret grows linearly
with $T$, and sublinear regret is impossible without imposing restrictions on the
time-varying model.
\end{lemma}

\begin{proof}
Let $D$ be the diameter of $\mathcal C$ and fix $c>0$ with $\eta c \le D/2$.
Consider any time $t$ and let $x_t\in\mathcal C$ be the iterate chosen by the algorithm
before observing $f_t$.
Since $\mathcal C$ is compact and convex with diameter $D$, we have
\[
R(x_t) := \sup_{z\in\mathcal C} \|z-x_t\| \ge \frac{D}{2} \ge \eta c.
\]
Thus there exists $z_t\in\mathcal C$ with $\|z_t-x_t\|\ge\eta c$.
Because $\mathcal C$ is convex, the segment $[x_t,z_t]\subseteq\mathcal C$, so there exists
$y_t\in[x_t,z_t]$ with $\|y_t-x_t\| = \eta c$.
Define
\[
s_t := \frac{1}{\eta}(x_t - y_t),
\]
so that $\|s_t\| = c$ and $y_t = x_t - \eta s_t \in \mathcal C$.

By the convexity and Lipschitz continuity of $g_0$ on $\mathcal C$, there exists
$u_t \in \partial g_0(y_t)$ with $\|u_t\| \le L_{g_0}$.
Since $0\in \mathcal{N}_{\mathcal C}(y_t)$, we also have
\[
u_t \in \partial g_0(y_t) + \mathcal{N}_{\mathcal C}(y_t) = \partial g(y_t).
\]
Set
\[
a_t := u_t - s_t,
\qquad\text{and}\qquad
f_t(x) := -\langle a_t,x\rangle.
\]
Then $\nabla f_t(x)\equiv -a_t$ for all $x$, so
$\|\nabla f_t(x)\| = \|a_t\| \le \|u_t\| + \|s_t\|\le L_{g_0} + c$ for all $t$.

Define $z_t := y_t + \eta u_t$.
By the proximal optimality condition for $g$,
\[
\frac{z_t-y_t}{\eta} = u_t \in \partial g(y_t)
\quad\Longrightarrow\quad
y_t = \prox_{\eta g}(z_t).
\]
Moreover,
\[
x_t - \eta\nabla f_t(x_t)
= x_t + \eta a_t
= x_t + \eta(u_t - s_t)
= (x_t - \eta s_t) + \eta u_t
= y_t + \eta u_t
= z_t.
\]
Therefore,
\[
\prox_{\eta g}\bigl(x_t - \eta\nabla f_t(x_t)\bigr)
= \prox_{\eta g}(z_t)
= y_t,
\]
and the proximal-gradient mapping satisfies
\[
\mathcal{P}^g_{\eta}\bigl(x_t;\nabla f_t(x_t)\bigr)
= \frac{1}{\eta}\Bigl(x_t - \prox_{\eta g}\bigl(x_t - \eta\nabla f_t(x_t)\bigr)\Bigr)
= \frac{1}{\eta}(x_t - y_t)
= s_t.
\]
Hence $\bigl\|\mathcal{P}^g_{\eta}\bigl(x_t;\nabla f_t(x_t)\bigr)\bigr\| = \|s_t\| = c$ for
every $t$, and summing over $t=1,\dots,T$ yields the claimed.
\end{proof}

\Cref{lem:necessity_variation_general} shows that sublinear proximal-gradient regret is
unattainable in general.
To obtain meaningful guarantees, additional structure must therefore be imposed on how
the losses evolve over time.
In what follows, we formalize this requirement through a bounded first-order variation
condition, which plays the role of a variability budget at the level of stationarity.

\subsection{The Variability Budget}
\label{sec:variability-budget}

In many online optimization models arising in operations research, the underlying objective
evolves over time due to changing costs, data, or system parameters.
Meaningful performance guarantees in such settings require that these changes be
sufficiently regular, a condition commonly formalized through stability or sensitivity
assumptions on the problem data or its solutions.
Such assumptions are standard in online optimization
\cite{CYLMLJZ12,RS13,BGZ15,JRSS15,HMC21,CCLS23}
and become indispensable in the nonconvex setting, where performance must be assessed
through stationarity rather than objective values.

\medskip

We adopt a \emph{gradient-based} notion of temporal regularity (see, e.g., \cite{CYLMLJZ12}), which is natural in
first-order oracle models and particularly suited to nonconvex analysis.
Rather than controlling function values, this budget directly bounds the evolution of
first-order optimality conditions over time.

\begin{definition}[Gradient-variation budget]
\label{def:budget}
The gradient-variation budget over a horizon $T$ is defined as
\begin{equation}\label{eq:V-def}
\mathcal{V}(T)
:= \sum_{t=1}^{T-1} \sup_{x\in\mathrm{dom} g}
\bigl\|\nabla f_{t+1}(x)-\nabla f_t(x)\bigr\|.
\end{equation}
\end{definition}
Since $\dom g$ is compact and each $\nabla f_t$ is continuous, $\mathcal{V}(T)$ is well defined.

From an optimization perspective, bounded gradient variation can be interpreted as a uniform sensitivity bound on first-order optimality conditions with respect to temporal perturbations of the problem data.
In many operations research models, underlying parameters—such as costs, demand forecasts, or risk estimates—evolve gradually rather than arbitrarily, reflecting conservative updates driven by physical, economic, or informational constraints.
Even in adversarial online formulations, variability budgets are commonly used to model environments that are nonstationary yet not fully unconstrained, thereby ruling out pathological sequences while retaining significant modeling flexibility.
A similar form of temporal regularity naturally arises in multiplayer games, where players’ payoffs change endogenously through the actions of others and therefore evolve in a coupled and typically smooth manner over time.

\begin{assumption}[Bounded gradient variation]
\label{ass:grad-var}
There exists $\alpha\in(0,1)$ such that
\[
\mathcal{V}(T) \le \mathcal{O}(T^{\alpha}).
\]
\end{assumption}

This assumption implies vanishing average variation, $\mathcal{V}(T)/T\to 0$, and is trivially
satisfied in the stationary (offline) setting where $f_t\equiv f$.

\medskip

The gradient-variation budget admits a direct interpretation,
capturing temporal variability either through infrequent regime shifts or through
small per-period parameter changes.
The following simple examples illustrate these two extremes.

\begin{example}[Linear objectives]
Let $f_t(x)=\langle c_t,x\rangle$ with $c_t\in\mathbb R^n$.
Then $\nabla f_t(x)=c_t$ and
\[
\mathcal V(T)=\sum_{t=1}^{T-1}\|c_{t+1}-c_t\|.
\]
Thus, bounded gradient variation amounts to controlling the cumulative variation of
the cost vector, a standard assumption in online linear optimization and dynamic
resource allocation.
\end{example}

\begin{example}[Online least squares]
Let $f_t(x)=\|\bA_t x-b_t\|^2$, where $\bA_t\in\mathbb R^{m\times n}$ and
$b_t\in\mathbb R^m$ vary over time.
If $\dom g$ is bounded, then
\[
\sup_{x\in\mathrm{dom} g}\|\nabla f_{t+1}(x)-\nabla f_t(x)\|
\;\lesssim\;
\|\bA_{t+1}^\top\bA_{t+1}-\bA_t^\top\bA_t\|
+
\|\bA_{t+1}^\top b_{t+1}-\bA_t^\top b_t\|.
\]
Accordingly, the gradient-variation budget controls the cumulative drift of the
empirical second- and first-order statistics of the model, as commonly encountered
in adaptive estimation with slowly varying systems.
\end{example}

\medskip

We next relate the gradient-variation budget to classical stability measures in
\emph{convex} optimization.
Bounded gradient variation controls function-value variation and, under strong
convexity, the path length of minimizers.
These relations are included only for context and are not used subsequently.

\begin{lemma}[Relation to function-variation budgets]
\label{lem:grad-to-func-var}
Fix $x_0\in \C$ and define the normalized losses $\tilde f_t(x)=f_t(x)-f_t(x_0)$.
Then
\[
\sum_{t=1}^{T-1}
\sup_{x\in \C}\big|\tilde f_{t+1}(x)-\tilde f_t(x)\big|
\;\le\;
D\,\mathcal{V}(T).
\]
\end{lemma}
\begin{proof}
	Follows from standard path-integration arguments using the fundamental theorem of calculus
and the compactness of $\C$.
\end{proof}

\begin{remark}[Relation to dynamic regret and variation budgets]
\label{rem:connection-besbes}
In the convex setting, our results are consistent with the classical dynamic-regret
landscape.
By \Cref{rem:dyn-to-local-to-regret}, local regret satisfies
$\Reg(T)\le \sqrt{T\,\mathrm{DReg}(T)}$, so the linear lower bound in
\Cref{lem:necessity_variation_general} implies that dynamic regret is necessarily linear
in the absence of temporal regularity, recovering the impossibility result underlying
\cite[Proposition~1]{BGZ15}.
Conversely, when the losses are convex, our gradient-variation budget implies a sublinear
function-variation budget via \Cref{lem:grad-to-func-var}, which is precisely the
regularity condition under which \cite{BGZ15} establish sublinear dynamic
regret.
Thus,  our framework provides a first-order extension that
naturally generalizes time-varying convex optimization to nonconvex composite objectives.
\end{remark}

\begin{lemma}[Relation to path-length budgets under strong convexity]
\label{lem:grad-to-pathlength}
Suppose that for each $t$, the composite objective $f_t+g$ is $\mu$-strongly convex on
$\C=\dom g$ for some $\mu>0$, and let $x_t^\star$ denote its unique minimizer.
Then
\[
\sum_{t=1}^{T-1}\|x_{t+1}^\star-x_t^\star\|
\;\le\;
\frac{1}{\mu}\,\mathcal{V}(T).
\]
\end{lemma}
\begin{proof}
This is a direct consequence of strong convexity and the strong monotonicity of the
subdifferential; see, e.g., \cite[Theorem 5.24]{B17}.
\end{proof}

\section{Methods and Guarantees}



\subsection{Online Proximal-Gradient Methods}
\label{sec:opg-framework}

We study online proximal-gradient methods that differ only in their access to first-order
information within each round.
Specifically, we consider two oracle-access regimes:
\emph{(i)} a nested (multi-call) variant that repeatedly applies proximal-gradient updates
until a prescribed stationarity tolerance is reached, and
\emph{(ii)} a single-call variant that performs exactly one proximal-gradient update per
round.

Both regimes are evaluated under the same local regret criterion.
Their distinction lies solely in oracle access, capturing a fundamental tradeoff between
stationarity tracking accuracy and per-round computational effort.
The nested method serves as a high-accuracy reference that closely tracks instantaneous
stationary points, while the single-call method represents a fully incremental online
update with minimal per-round cost.

\medskip
\noindent
\begin{algorithm}[ht]
\caption{Online Nested Proximal-Gradient Method}
\label{alg:4}
\KwIn{$x_1\in\mathbb{R}^n$, $\{\delta_t\}_{t\in[T]}\subset\mathbb{R}_{>0}$, and $\eta>0$.}
\For{$t=1,2,\ldots,T$}{
    $f_t:\mathbb{R}^n\to\mathbb{R}$ is revealed\;
    $\by_t^1 \leftarrow x_t$, \quad $k\leftarrow1$\;
    \While{$\bigl\|\mathcal{P}^g_{\eta}(\by_t^k;\nabla f_t(\by_t^k))\bigr\|>\delta_t$}{
        $\by_t^{k+1}
        \leftarrow
        \prox_{\eta g}\!\left(\by_t^k-\eta\nabla f_t(\by_t^k)\right)$\;
        $k\leftarrow k+1$\;
    }
    $x_{t+1}\leftarrow \by_t^k$\;
}
\end{algorithm}

\medskip
\noindent
\begin{algorithm}[t]
\caption{Single-call Proximal Gradient (SC-PG)}
\label{alg:scpg}
\KwIn{$x_1\in\dom g$, stepsize $\eta\in(0,1/L_f]$.}
\For{$t=1,2,\ldots,T$}{
    Obtain one oracle call $\nabla f_t(x_t)$\;
    $x_{t+1}
    \leftarrow
    \prox_{\eta g}\bigl(x_t-\eta \nabla f_t(x_t)\bigr)$\;
}
\end{algorithm}

We analyze these two oracle-access regimes separately in the sequel.

\subsection{Nested Proximal-Gradient Method}
\label{sec:nested-pg}

We first analyze the nested proximal-gradient method, which allows repeated oracle access
to the same loss function within each round.
This oracle model serves as an idealized benchmark for isolating the effect of accurately
solving each instantaneous problem on the resulting local regret.

\paragraph{Oracle model.}
After observing $f_t$, the algorithm may query $\nabla f_t(\cdot)$ multiple times in order
to approximately solve the instantaneous composite problem $f_t+g$ to a prescribed
stationarity tolerance.

\begin{remark}[Repeated oracle access in data-driven models]
While the nested oracle model allows multiple gradient queries per round, this setting is
natural in several data-driven online optimization problems where the underlying loss
function remains structurally fixed and only the data evolve over time.
Examples include portfolio optimization with time-varying returns and risk estimates, or regression models under nonstationary data streams, where the same objective function is
queried at multiple points as new information arrives.
In such settings, repeated gradient evaluations within a round correspond to accessing
the same model with updated data, rather than solving an offline problem.

\end{remark}

\medskip
The following assumption specifies how accurately each instantaneous problem is solved.

\begin{assumption}[Diminishing local accuracy]
\label{ass:diminishing-accuracy}
The sequence $\{\delta_t\}_{t=1}^T$ is nonincreasing and satisfies, for some
$\alpha\in[0,1)$,
\[
\sum_{t=1}^T \delta_t = O(T^\alpha).
\]
\end{assumption}

This condition ensures that the cumulative stationarity error incurred by inexact
per-round solutions grows sublinearly with the time horizon.
It accommodates standard accuracy schedules such as $\delta_t=t^{-1/2}$ (yielding
$\alpha=\tfrac12$) or $\delta_t=T^{\alpha-1}$ when the horizon is known in advance.

\medskip
We now establish a bound on the local regret of \Cref{alg:4}.

\begin{theorem}[Local regret of the nested method]
\label{thm:nested-local-regret}
Suppose that \Cref{ass:diminishing-accuracy} holds.
Then the iterates $\{x_t\}_{t=1}^T$ generated by \Cref{alg:4} satisfy
\[
\reg(T)
\;\le\;
\frac{D}{\eta}
\;+\;
\sum_{t=1}^{T-1} \delta_t
\;+\;
\mathcal{V}(T).
\]
In particular, if $\mathcal{V}(T)=O(T^\alpha)$, then $\reg(T)=O(T^\alpha)$.
\end{theorem}

\begin{proof}
By definition of the local regret,
\[
\reg(T)
=
\sum_{t=1}^T
\bigl\|\mathcal{P}^g_\eta(x_t;\nabla f_t(x_t))\bigr\|.
\]

For $t\ge2$, using the Lipschitz continuity of the proximal-gradient mapping
with respect to the gradient argument (cf. \Cref{lem:T-eta-sensitivity}),
\[
\bigl\|\mathcal{P}^g_\eta(x_t;\nabla f_t(x_t))\bigr\|
\le
\bigl\|\mathcal{P}^g_\eta(x_t;\nabla f_{t-1}(x_t))\bigr\|
+
\|\nabla f_t(x_t)-\nabla f_{t-1}(x_t)\|.
\]

By the stopping rule of the inner loop at time $t-1$,
\[
\bigl\|\mathcal{P}^g_\eta(x_t;\nabla f_{t-1}(x_t))\bigr\|\le \delta_{t-1}.
\]
Moreover,
\[
\|\nabla f_t(x_t)-\nabla f_{t-1}(x_t)\|
\le
\sup_{x\in\dom g}\|\nabla f_t(x)-\nabla f_{t-1}(x)\|.
\]
The right-hand side is precisely the per-round gradient variation term appearing in the
variation budget $\mathcal{V}(T)$.

For $t=1$, since both $x_1$ and $\prox_{\eta g}(x_1-\eta\nabla f_1(x_1))$ belong to
$\dom g$, we have
$\|\mathcal{P}^g_\eta(x_1;\nabla f_1(x_1))\|\le D/\eta$.

Summing over $t=1,\ldots,T$ yields the stated bound.
\end{proof}

This bound shows that the nested method tracks the time-varying stationary set up to the
combined effects of temporal variation, quantified by $\mathcal{V}(T)$, and inexact inner
solutions, quantified by $\sum_{t=1}^{T-1}\delta_t$.

\medskip

When the losses are convex, the local regret bound immediately yields a dynamic
regret guarantee via \Cref{thm:regret-local-to-dynamic}.

\begin{corollary}[Dynamic regret under convexity]
\label{cor:nested-dynamic-regret}
Suppose that \Cref{ass:diminishing-accuracy} holds and that each $f_t$ is convex.
Then
\[
\DReg(T)
\;\le\;
C\,\reg(T),
\]
where $C>0$ is the constant from \Cref{thm:regret-local-to-dynamic}.
In particular, if the variation budget satisfies $\mathcal{V}(T)=O(T^\alpha)$ for some
$\alpha\in[0,1)$, then both the local regret and the dynamic regret obey
$\reg(T)=O(T^\alpha)$ and $\DReg(T)=O(T^\alpha)$, and hence $\DReg(T)=o(T)$.
\end{corollary}

We now bound the total number of inner proximal-gradient iterations.
Let $\tau_t$ denote the number of inner-loop iterations executed at time $t$,
and set $\tau=\sum_{t=1}^T \tau_t$.

\begin{theorem}[Iteration complexity of the nested method]
\label{thm:complexity-det}
Suppose that \Cref{ass:diminishing-accuracy} holds true.
Then for each $t\in[T]$,
\[
\tau_t
\;\le\;
1 + \frac{4M}{\eta(1-\eta L_f)\,\delta_t^2},
\]
and consequently
\[
\tau
\;\le\;
T + \sum_{t=1}^T \frac{4M}{\eta(1-\eta L_f)\,\delta_t^2}.
\]
\end{theorem}

\begin{proof}
By the standard sufficient decrease property of the proximal-gradient method
(cf.\ \Cref{cor:pg-ineq}), for each inner iteration $k$ at round $t$ we have
\[
F_t(\by_t^k)-F_t(\by_t^{k+1})
\;\ge\;
\frac{\eta(1-\eta L_f)}{2}
\bigl\|\mathcal{P}^g_\eta(\by_t^k;\nabla f_t(\by_t^k))\bigr\|^2.
\]
Summing over $k=1,\ldots,\tau_t-1$ and using the stopping rule
$\|\mathcal{P}^g_\eta(\by_t^k;\nabla f_t(\by_t^k))\|>\delta_t$ for all such $k$
yields
\[
(\tau_t-1)\,\frac{\eta(1-\eta L_f)}{2}\,\delta_t^2
\;\le\;
F_t(\by_t^1)-F_t(\by_t^{\tau_t}).
\]
By the boundedness assumption $|F_t(x)|\le M$ on $\dom g$, we have
$F_t(\by_t^1)\le M$ and $F_t(\by_t^{\tau_t})\ge -M$, so
\[
F_t(\by_t^1)-F_t(\by_t^{\tau_t}) \;\le\; 2M.
\]
Combining the two inequalities gives
\[
(\tau_t-1)\,\frac{\eta(1-\eta L_f)}{2}\,\delta_t^2
\;\le\;
2M,
\]
which implies
\[
\tau_t
\;\le\;
1 + \frac{4M}{\eta(1-\eta L_f)\,\delta_t^2}.
\]
Summing over $t=1,\ldots,T$ yields the stated bound on $\tau$.
\end{proof}

To illustrate the tradeoff between local-regret performance and inner-loop complexity
captured by \Cref{thm:nested-local-regret} and \Cref{thm:complexity-det}, we consider the
following representative parameter choice.

\paragraph{Example parameter choice.}
If $\delta_t=T^{\alpha-1}$ and $\eta=1/(2L_f)$, then
\[
\reg(T)=O(T^\alpha),
\qquad
\tau=O(T^{3-2\alpha}).
\]
In particular, choosing $\alpha=\tfrac12$ yields $\reg(T)=O(\sqrt{T})$ and
$\tau=O(T^2)$.
The $\sqrt{T}$ regret rate is consistent with classical dynamic-regret bounds
under variation-budget assumptions, while the quadratic complexity reflects
the standard $O(\delta^{-2})$ cost of achieving first-order stationarity via
proximal-gradient methods.

\medskip
The nested method leverages repeated oracle access to closely track instantaneous
stationary points, but may incur substantial per-round computational cost.
We now examine the opposite extreme, where updates are fully incremental and oracle access
is severely limited.

\subsection{Single-call Proximal-Gradient Method}
\label{sec:single-call-pg}

We consider an online oracle model in which exactly one first-order oracle call is permitted
per round.
In this regime, the decision is updated by a single proximal-gradient step with respect to
the current loss, yielding a minimal-cost update rule.
Relative to the nested method, this restriction fundamentally limits the accuracy with which
the evolving stationary set can be tracked, and leads to weaker, but still meaningful, local
regret guarantees.

\begin{theorem}[Local regret of the single-call method]
\label{thm:scpg-local-regret}
Let $\{x_t\}_{t=1}^{T+1}$ be generated by \Cref{alg:scpg} with stepsize
$\eta\in(0,1/L_f)$.
Then, for any $T\ge1$,
\begin{equation}\label{eq:scpg-squared}
\sum_{t=1}^{T}
\Bigl\|
\mathcal{P}^g_{\eta}\bigl(x_t;\nabla f_t(x_t)\bigr)
\Bigr\|^2
\;\le\;
\frac{2}{\eta(1-\eta L_f)}\Bigl(4M + D\,\mathcal{V}(T)\Bigr).
\end{equation}
Consequently, the local regret \eqref{eq:local regret} satisfies
\begin{equation}\label{eq:scpg-regret}
\reg(T)
\;\le\;
\sqrt{T\sum_{t=1}^{T}
\Bigl\|
\mathcal{P}^g_{\eta}\bigl(x_t;\nabla f_t(x_t)\bigr)
\Bigr\|^2}
\;\le\;
\sqrt{\frac{2T}{\eta(1-\eta L_f)}\Bigl(4M + D\,\mathcal{V}(T)\Bigr)}.
\end{equation}
In particular, under \Cref{ass:grad-var} with $\mathcal{V}(T)=\mathcal{O}(T^\alpha)$ for some
$\alpha\in(0,1)$, we obtain the sublinear bound
\[
\reg(T)=\mathcal{O}\bigl(T^{(1+\alpha)/2}\bigr).
\]
\end{theorem}
\begin{proof}
Fix any anchor point $x^\circ\in\mathcal{C}$ and define the shifted losses
\[
\tilde f_t(x):= f_t(x)-f_t(x^\circ),
\qquad
\tilde F_t(x):=\tilde f_t(x)+g(x).
\]
This shift does not change gradients, i.e., $\nabla \tilde f_t \equiv \nabla f_t$,
hence \Cref{alg:scpg} generates the same iterates when $f_t$ is replaced by $\tilde f_t$.
Moreover, since $|f_t(x)|\le M$ and $|F_t(x)|\le M$ for all $x\in\mathcal{C}$ and $t\in[T]$,
we have $|\tilde f_t(x)|\le 2M$ and $|\tilde F_t(x)|\le 2M$ on $\mathcal{C}$.

\smallskip
\noindent\textbf{Step 1: One-step decrease and a squared-residual bound.}
For $t=1,\dots,T$, define
\[
G_t
\;:=\;
\mathcal{P}_{\eta}^{g}\bigl(x_t;\nabla f_t(x_t)\bigr)
\;=\;
\frac{1}{\eta}\bigl(x_t-x_{t+1}\bigr),
\]
where the last equality follows from the update rule in \Cref{alg:scpg}
and the definition \eqref{eq:pg-residual}.
Applying \Cref{lemma:sufficient decrease} to $f=f_t$, $g=g$, and $x=x_t$ gives
\[
F_t(x_{t+1})
\le
F_t(x_t)
-
\frac{1-\eta L_f}{2\eta}\|x_{t+1}-x_t\|^2.
\]
Since $\|x_{t+1}-x_t\|^2=\eta^2\|G_t\|^2$, we obtain
\begin{equation}\label{eq:pg-descent-scpg-correct}
F_t(x_{t+1})
\;\le\;
F_t(x_t)\;-\;\frac{\eta(1-\eta L_f)}{2}\,\|G_t\|^2.
\end{equation}
As the shift from $F_t$ to $\tilde F_t$ is constant in $x$, the same inequality holds for
$\tilde F_t$:
\[
\tilde F_t(x_{t+1})
\;\le\;
\tilde F_t(x_t)\;-\;\frac{\eta(1-\eta L_f)}{2}\,\|G_t\|^2,
\]
and hence
\begin{equation}\label{eq:sq-bound-basic-scpg-correct}
\|G_t\|^2
\;\le\;
\frac{2}{\eta(1-\eta L_f)}
\Bigl(\tilde F_t(x_t)-\tilde F_t(x_{t+1})\Bigr).
\end{equation}
Summing \eqref{eq:sq-bound-basic-scpg-correct} over $t=1,\dots,T$ yields
\begin{equation}\label{eq:sumsq-scpg-correct}
\sum_{t=1}^{T} \|G_t\|^2
\;\le\;
\frac{2}{\eta(1-\eta L_f)}
\sum_{t=1}^{T}
\Bigl(\tilde F_t(x_t)-\tilde F_t(x_{t+1})\Bigr).
\end{equation}

\smallskip
\noindent\textbf{Step 2: Controlling time variation via $\mathcal{V}(T)$.}
For each $t=1,\dots,T-1$, decompose
\[
\tilde F_t(x_t)-\tilde F_t(x_{t+1})
=
\bigl(\tilde F_t(x_t)-\tilde F_{t+1}(x_{t+1})\bigr)
+
\bigl(\tilde F_{t+1}(x_{t+1})-\tilde F_t(x_{t+1})\bigr).
\]
Summing this identity over $t=1,\dots,T-1$ and then adding the remaining term
$\tilde F_T(x_T)-\tilde F_T(x_{T+1})$ gives
\begin{align}
\sum_{t=1}^{T}\bigl(\tilde F_t(x_t)-\tilde F_t(x_{t+1})\bigr)
&=
\underbrace{\sum_{t=1}^{T-1}\bigl(\tilde F_t(x_t)-\tilde F_{t+1}(x_{t+1})\bigr)
+\bigl(\tilde F_T(x_T)-\tilde F_T(x_{T+1})\bigr)}_{(\mathrm{I})}
\nonumber\\
&\qquad+
\underbrace{\sum_{t=1}^{T-1}\bigl(\tilde F_{t+1}(x_{t+1})-\tilde F_t(x_{t+1})\bigr)}_{(\mathrm{II})}.
\label{eq:decomp-scpg}
\end{align}
The bracketed term $(\mathrm{I})$ telescopes:
\[
(\mathrm{I})
=
\tilde F_1(x_1)-\tilde F_T(x_T)+\tilde F_T(x_T)-\tilde F_T(x_{T+1})
=
\tilde F_1(x_1)-\tilde F_T(x_{T+1})
\;\le\; 4M,
\]
where we used $|\tilde F_t(x)|\le 2M$ for all $x\in\mathcal{C}$ and $t\in[T]$.

For term $(\mathrm{II})$, note that $g$ cancels, and define
$h_t(x):=\tilde f_{t+1}(x)-\tilde f_t(x)$.
Then
\[
\nabla h_t(x)=\nabla f_{t+1}(x)-\nabla f_t(x),
\qquad
\|\nabla h_t(x)\|
\le
v_t
:=
\sup_{z\in\mathcal{C}}
\|\nabla f_{t+1}(z)-\nabla f_t(z)\|,
\]
where by definition, $\mathcal{V}(T)=\sum_{t=1}^{T-1} v_t$.
Thus $h_t$ is $v_t$-Lipschitz on the convex set $\mathcal{C}$.
Since $h_t(x^\circ)=0$, for any $x\in\mathcal{C}$,
\[
|h_t(x)|
\le
v_t\|x-x^\circ\|
\le
D\,v_t.
\]
In particular,
\[
\tilde F_{t+1}(x_{t+1})-\tilde F_t(x_{t+1})
=
h_t(x_{t+1})
\le
D\,v_t.
\]
Summing over $t=1,\dots,T-1$ yields for the bracketed term $(\mathrm{II})$
\[
\sum_{t=1}^{T-1}\bigl(\tilde F_{t+1}(x_{t+1})-\tilde F_t(x_{t+1})\bigr)
\le
D\sum_{t=1}^{T-1} v_t
=
D\,\mathcal{V}(T).
\]

Combining the bounds on $(\mathrm{I})$ and $(\mathrm{II})$ in \eqref{eq:decomp-scpg} gives
\[
\sum_{t=1}^{T}\bigl(\tilde F_t(x_t)-\tilde F_t(x_{t+1})\bigr)
\;\le\;
4M + D\,\mathcal{V}(T).
\]
Substituting into \eqref{eq:sumsq-scpg-correct} yields \eqref{eq:scpg-squared}.

\smallskip
\noindent\textbf{Step 3: From squared bound to local regret.}
By the Cauchy--Schwarz inequality,
\[
\reg(T)
=
\sum_{t=1}^{T} \|G_t\|
\;\le\;
\sqrt{T\sum_{t=1}^{T}\|G_t\|^2}.
\]
Substituting the bound \eqref{eq:scpg-squared} yields \eqref{eq:scpg-regret}.
Finally, under \Cref{ass:grad-var} with $\mathcal{V}(T)=\mathcal{O}(T^\alpha)$,
we obtain $\reg(T)=\mathcal{O}(T^{(1+\alpha)/2})$.
\end{proof}

When the losses are convex, the local regret bound of
\Cref{thm:scpg-local-regret} can be converted into a dynamic regret guarantee
via the local-to-dynamic reduction established in \Cref{thm:regret-local-to-dynamic}.

\begin{corollary}[Dynamic regret under convexity]
\label{cor:scpg-dynamic-regret}
Assume in addition to the standing assumptions that each $f_t$ is convex on $\mathcal{C}$.
Let $\{x_t\}_{t=1}^{T}$ be generated by \Cref{alg:scpg} with a constant stepsize
$\eta\in(0,1/L_f)$, and let $\{x_t^\star\}_{t=1}^T$ be any minimizer path with
$x_t^\star\in\argmin_{x\in\mathcal{C}} F_t(x)$.
Then the dynamic regret
is bounded as
\begin{equation}\label{eq:scpg-dyn-regret}
\DReg(T)
\;\le\;
\left(D + \frac{M+L_g}{L_f}\right)\reg(T).
\end{equation}
Consequently, from \Cref{thm:scpg-local-regret},
we obtain
\begin{equation}\label{eq:scpg-dyn-regret-rate}
\DReg(T)
\;\le\;
\left(D + \frac{M+L_g}{L_f}\right)
\sqrt{\frac{2T}{\eta(1-\eta L_f)}\Bigl(4M + D\,\mathcal{V}(T)\Bigr)}.
\end{equation}
In particular, if $\mathcal{V}(T)=\mathcal{O}(T^\alpha)$ for some $\alpha\in(0,1)$, then
$\DReg(T)=\mathcal{O}\!\left(T^{(1+\alpha)/2}\right)$.
\end{corollary}

\begin{proof}
The bound \eqref{eq:scpg-dyn-regret} follows by applying
\Cref{thm:regret-local-to-dynamic} 
to the decision sequence $\{x_t\}_{t=1}^T$ produced by \Cref{alg:scpg}.
Substituting the local-regret bound of \Cref{thm:scpg-local-regret} yields
\eqref{eq:scpg-dyn-regret-rate}.
\end{proof}

We turn to demonstrate how the proposed framework specializes naturally to nonconvex  multiplayer games.

\section{Special Case: Nonconvex Multiplayer Games}
\label{sec:case 1 multiplayer}

Game-theoretic models provide a natural testbed for online nonconvex optimization, as players repeatedly react to a nonstationary environment induced by the evolving actions of others;
see, e.g., \cite{FC97,PR16,RHLNSH20,BR20,HM20,LXL23,CT24}.

We consider a nonconvex, nonsmooth multiplayer game with $K$ players, where the aggregated
decision vector is
\[
x = (x^1,x^2,\ldots,x^K) \in \mathcal{C}
:= \mathcal{C}^1 \times \cdots \times \mathcal{C}^K,
\]
with each $\mathcal{C}^i \subseteq \mathbb{R}^{n_i}$ compact and convex.
The joint composite objective is
\begin{equation}\label{eq:multiplayers}
F(x)
=
H(x^1,x^2,\ldots,x^K)
+
\sum_{i=1}^K g^i(x^i),
\end{equation}
where $H$ is possibly nonconvex and smooth, and each $g^i$ is proper, closed, and convex.

Following \cite{PR16}, at each time $t$ player $i$ observes the current actions
$x_t^{-i}$ of the other players and faces the individual optimization problem
\begin{equation}\label{eq:two-players}
F_t^i(z)
=
H(z,x_t^{-i}) + g^i(z)
\;\equiv\;
H_t^i(z) + g^i(z),
\end{equation}
where the time dependence arises endogenously through the evolving joint action profile
$\{x_t\}_{t\ge1}$.
Hence, from the perspective of each player, \eqref{eq:two-players} defines an online
nonconvex composite optimization problem with time-varying losses driven by the dynamics
of the game.

Thus, for each player $i$, the function $H_t^i$ plays the role of the time-varying smooth loss
$f_t$, while $g^i$ plays the role of the regularizer $g$ in the online composite model
\eqref{prob:1}.
All standing assumptions imposed on $f_t$ and $g$ are assumed to hold uniformly on each
player’s feasible set.

In this setting, the local regret of player~$i$,
\[
\reg_i(T)
=
\sum_{t=1}^T
\left\|
\mathcal{P}^{g^i}_{\eta}
\bigl(x_t^i;\nabla_{x_i}H(x_t)\bigr)
\right\|,
\]
quantifies the cumulative violation of first-order stationarity of the time-varying
objective $F_t^i$.

Consequently, small local regret implies approximate first-order optimality of each
player’s actions against the realized sequence of opponents’ strategies.
In particular, $\varepsilon$-local regret corresponds to an $\varepsilon$-approximate
first-order Nash equilibrium (FNE) in the sense of \cite{BR20}, and is closely related to
the quasi-Nash equilibrium (QNE) concept of \cite{PR16}.
Existence of a (quasi-)Nash equilibrium is ensured under standard regularity conditions on
$H$, such as twice continuous differentiability; see \cite[Proposition~4.2]{PR16}.

Throughout this section, we assume that each player applies either
\Cref{alg:4} or \Cref{alg:scpg} to determine her next action.

\medskip 

To invoke the regret guarantees developed in previous sections, we require a bound on the
temporal variation of each player’s effective loss (see \cite{Zinkevich03} for a similar assumption).

\begin{lemma}[Variability budget induced by joint play]
\label{lem:game-variation}
Suppose the sequence of joint actions $\{x_t\}_{t=1}^T$ satisfies
\[
\sum_{t=1}^{T-1}\|x_{t+1}-x_t\| \;\le\; C\,T^\alpha
\]
for some $C>0$ and $\alpha\in[0,1)$, and that $\nabla H$ is Lipschitz continuous with
respect to all coordinates with constant $L_H$.
Then \Cref{ass:grad-var} holds for each player $i$, with
$\mathcal{V}_i(T)=\mathcal{O}(T^\alpha)$.
\end{lemma}

\begin{proof}
Fix a player $i$ and $t\in\{1,\dots,T-1\}$.
For any $u\in\dom g^i$, we have
\[
\nabla_{x_i}H_t^i(u)
=
\nabla_{x_i}H(u,x_t^{-i}),
\qquad
\nabla_{x_i}H_{t+1}^i(u)
=
\nabla_{x_i}H(u,x_{t+1}^{-i}).
\]
By $L_H$-Lipschitz continuity of $\nabla H$ with respect to all coordinates,
\[
\|\nabla_{x_i}H_t^i(u)-\nabla_{x_i}H_{t+1}^i(u)\|
=
\|\nabla_{x_i}H(u,x_t^{-i})-\nabla_{x_i}H(u,x_{t+1}^{-i})\|
\le
L_H\|x_t^{-i}-x_{t+1}^{-i}\|.
\]
Taking the supremum over $u\in\dom g^i$ yields
\[
v_t^i
\;:=\;
\sup_{u\in\dom g^i}
\|\nabla_{x_i}H_t^i(u)-\nabla_{x_i}H_{t+1}^i(u)\|
\;\le\;
L_H\|x_t^{-i}-x_{t+1}^{-i}\|
\;\le\;
L_H\|x_{t+1}-x_t\|,
\]
where the last inequality follows from $x_t=(x_t^i,x_t^{-i})$.
Summing over $t=1,\dots,T-1$ and using the assumed bound on joint variation, we obtain
\[
\mathcal{V}_i(T)
=
\sum_{t=1}^{T-1} v_t^i
\;\le\;
L_H\sum_{t=1}^{T-1}\|x_{t+1}-x_t\|
\;\le\;
L_H C\,T^\alpha,
\]
so \Cref{ass:grad-var} holds for player $i$ with $\mathcal{V}_i(T)=\mathcal{O}(T^\alpha)$.
\end{proof}

The assumption in \Cref{lem:game-variation} captures regimes in which players adapt to one another in a controlled
manner, in the sense that the cumulative variation of the joint action profile grows
sublinearly with the time horizon, leading to a gradually evolving effective environment
for each player.

We now state the resulting equilibrium guarantee.

\begin{corollary}[Approximate first-order Nash equilibria under proximal-gradient play]
\label{cor:fne-unified}
Consider the multiplayer game defined in \eqref{eq:multiplayers}–\eqref{eq:two-players},
and suppose that the gradient of $H$ is Lipschitz continuous in all coordinates.
Assume that the joint action sequence $\{x_t\}_{t=1}^T$ satisfies
\[
\sum_{t=1}^{T-1}\|x_{t+1}-x_t\| \;\le\; C\,T^\alpha
\qquad\text{for some }C>0\text{ and }\alpha\in(0,1).
\]

Then the following statements hold.

\begin{enumerate}
\item \textbf{Multi-call (nested) updates.}
If each player applies the nested proximal-gradient method \Cref{alg:4} with a fixed
stepsize $\eta\in(0,1/L_f)$ and diminishing stationarity tolerances
$\{\delta_t\}$ satisfying \Cref{ass:diminishing-accuracy}, then
\[
\min_{t\in[T]}
\sum_{i=1}^K
\left\|
\mathcal{P}^{g^i}_{\eta}
\bigl(x_t^i;\nabla_{x_i}H(x_t)\bigr)
\right\|
\;=\;
O\!\left(K\,T^{\alpha-1}\right).
\]
Consequently, an $\varepsilon$-approximate first-order Nash equilibrium (in the sense that
the above residual is at most $\varepsilon$ for some $t\le T$) is attained after
\[
T
=
O\!\left(\left(\frac{K}{\varepsilon}\right)^{\frac{1}{1-\alpha}}\right)
\]
iterations.

\item \textbf{Single-call updates.}
If each player applies the single-call proximal-gradient method \Cref{alg:scpg} with a fixed
stepsize $\eta\in(0,1/L_f)$, then
\[
\min_{t\in[T]}
\sum_{i=1}^K
\left\|
\mathcal{P}^{g^i}_{\eta}
\bigl(x_t^i;\nabla_{x_i}H(x_t)\bigr)
\right\|
\;=\;
O\!\left(K\,T^{\frac{\alpha-1}{2}}\right).
\]
Consequently, an $\varepsilon$-approximate first-order Nash equilibrium is attained after
\[
T
=
O\!\left(
\left(\frac{K}{\varepsilon}\right)^{\frac{2}{1-\alpha}}
\right)
\]
iterations.
\end{enumerate}

In both cases, if the game admits at least one first-order Nash equilibrium, then any
accumulation point of a subsequence along which the averaged stationarity residual
vanishes satisfies the first-order Nash conditions.
\end{corollary}

\begin{proof}
By \Cref{lem:game-variation}, the path-length condition implies that each player’s effective
loss satisfies the gradient-variation condition with
$\mathcal{V}_i(T)=O(T^\alpha)$.

\smallskip
\noindent\emph{Multi-call updates.}
Applying the local-regret bound of \Cref{thm:nested-local-regret} to each player yields
$\reg_i(T)=O(T^\alpha)$.
Summing over players and averaging over time gives
\[
\min_{t\in[T]}
\sum_{i=1}^K
\left\|
\mathcal{P}^{g^i}_{\eta}
\bigl(x_t^i;\nabla_{x_i}H(x_t)\bigr)
\right\|
\;\le\;
\frac{1}{T}\sum_{i=1}^K \reg_i(T)
=
O\!\left(K\,T^{\alpha-1}\right),
\]
which implies the stated rate.

\smallskip
\noindent\emph{Single-call updates.}
Applying \Cref{thm:scpg-local-regret} player-wise yields
$\reg_i(T)=O(T^{(1+\alpha)/2})$ for all $i$.
Summing over players and averaging over time gives
\[
\min_{t\in[T]}
\sum_{i=1}^K
\left\|
\mathcal{P}^{g^i}_{\eta}
\bigl(x_t^i;\nabla_{x_i}H(x_t)\bigr)
\right\|
=
O\!\left(K\,T^{\frac{\alpha-1}{2}}\right).
\]

\smallskip
If first-order Nash equilibria exist, standard compactness arguments imply that any sequence
with vanishing averaged stationarity residual admits accumulation points that satisfy the
first-order Nash conditions.
\end{proof}

This case study illustrates how the proposed  framework applies naturally to nonconvex multiplayer games, where each player faces an endogenous nonstationary and nonconvex optimization problem induced by the evolving actions of others.
The local regret guarantees derived in \Cref{sec:opg-framework} translate directly
into quantitative bounds on first-order Nash equilibrium violations under both nested and
single-call oracle models.
The resulting tradeoffs between tracking accuracy, oracle access, and temporal variation highlight fundamental limitations of incremental play in nonstationary environments,
while requiring no potential structure, convexity of the joint objective, or convergence to equilibrium.

\section{Numerical Illustrations}
\label{sec:Numerical}

We present a small set of numerical experiments illustrating the qualitative behavior
of the online proximal-gradient framework in representative nonconvex online settings.
The goal is to demonstrate phenomena predicted by the theory—stability under nonstationarity
and meaningful stationarity tracking—rather than to validate asymptotic rates or compare
algorithms. All experiments focus on the nested oracle regime.

\subsection{Nonconvex Multiplayer Games}
\label{sec:multiplayer numeric}

We first illustrate the behavior of \Cref{alg:4} in the multiplayer setting
introduced in \Cref{sec:case 1 multiplayer}.
This experiment serves as a concrete instantiation of the interpretation of local regret as a
measure of approximate first-order Nash equilibrium in nonconvex games.

We consider a synthetic nonconvex game motivated by a single-commodity market (e.g.,
electricity), consisting of one large producer and several small producers, each capable of
producing the commodity using multiple sources.
The joint decision variable is
$\mathbf{x}=(x_1,x_2,\ldots,x_N)\in\mathbb{R}^{nN}$, where $x_i\in\mathbb{R}^n$ denotes
the production levels of player $i$ across $n$ sources.
The total production
$
Q=\sum_{i\in[N]}\sum_{j\in[n]} x_{i,j}
$
determines the market price through the demand curve
\[
p(Q)=\frac{p_{\max}}{1+e^{kQ}},
\]
with $p_{\max}>0$ and $k\ge0$ controlling price sensitivity.

Player heterogeneity is modeled through individual cost functions
\[
\mathrm{cost}_i(x_i)
=
\log\!\bigl(1+\|w_i\circ x_i\|_2^2\bigr)
+
\lambda\|x_i\|_2^2,
\]
which induce nonconvex objectives.
Each player $i\in[N]$ solves
\[
\min_{x_i\in \C_i}
\;F_i(x_i,x_{-i})
=
\mathrm{cost}_i(x_i)
-
p(Q)\sum_{j\in[n]} x_{i,j},
\]
where $C_i=[0,r_{\text{small}}]^n$ for small producers and
$C_i=[0,r_{\text{large}}]^n$ for the large producer.

All players apply \Cref{alg:4} as described in \Cref{sec:case 1 multiplayer}, initializing from
the zero vector and using a common constant step size.
To illustrate robustness beyond the idealized setting, we additionally consider perturbed
gradients with zero-mean Gaussian noise whose first moment decays over time.
 This noise model is not covered by the theory and is included solely as a stress test to
illustrate qualitative robustness; no theoretical guarantees are claimed in this setting.

For reference only, we also compute a centralized benchmark corresponding to a single decision
maker minimizing the aggregate objective
\[
\min_{\mathbf{x}}
\left\{
\sum_{i\in[N]}\mathrm{cost}_i(x_i)
-
p(Q)\,Q
\right\},
\]
which is solved offline using SciPy's \texttt{L-BFGS-B} solver and reported as an average
per-player value.
This benchmark is \emph{not} a Nash equilibrium and is included solely to provide qualitative
context.

\begin{figure}[ht]
	\centering
	\begin{subfigure}[b]{0.45\textwidth}
		\centering
		\includegraphics[width=\textwidth]{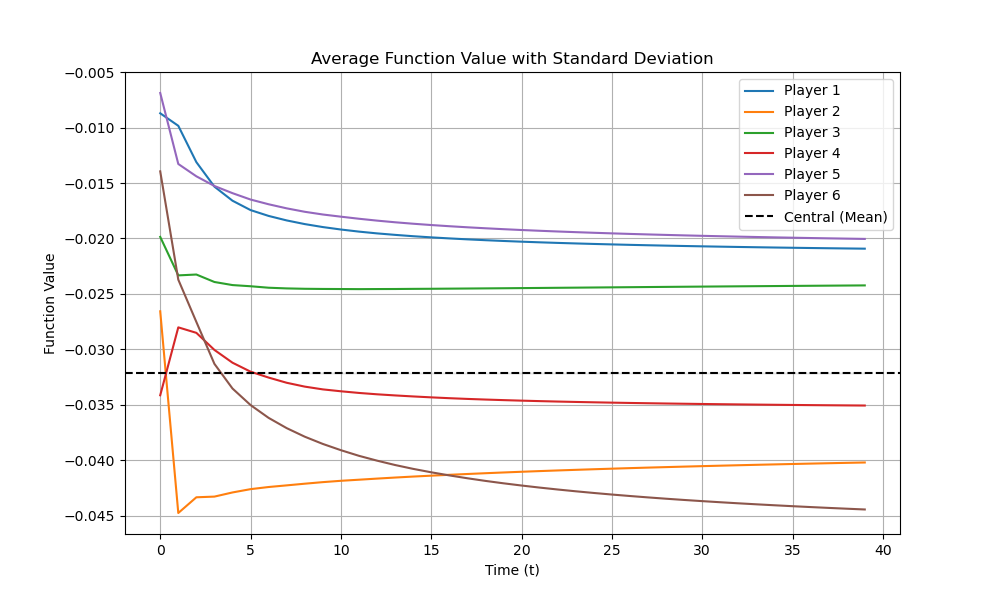}
		\caption{Single run.}
	\end{subfigure}
	\hfill
	\begin{subfigure}[b]{0.45\textwidth}
		\centering
		\includegraphics[width=\textwidth]{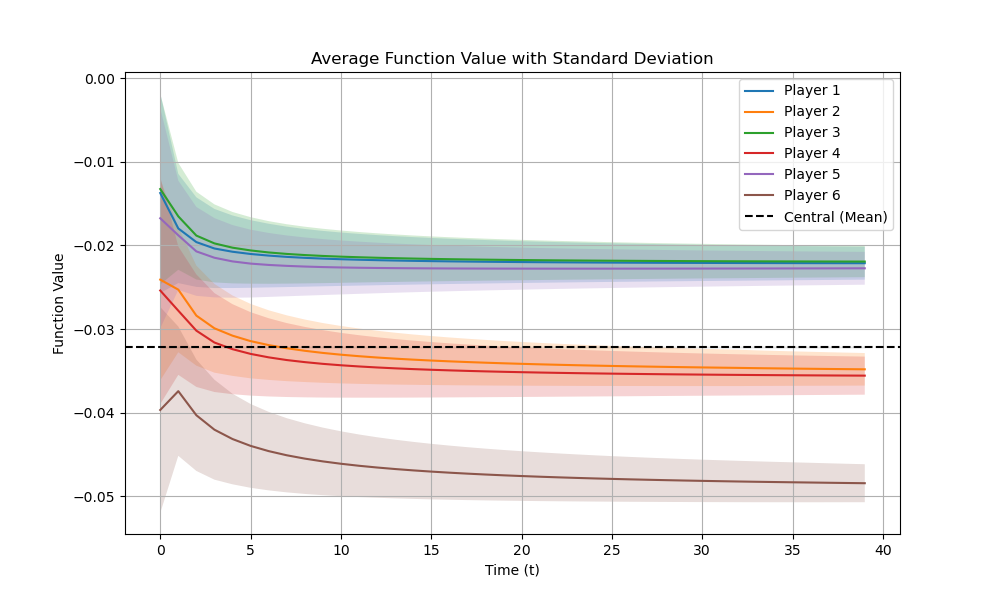}
		\caption{Average over 50 runs.}
	\end{subfigure}
	\caption{Per-player objective values in a nonconvex multiplayer market under nested proximal-gradient
updates.
}
	\label{fig:MP}
\end{figure}

The results shown in \Cref{fig:MP} correspond to a market with $N=6$ producers (one large and
five small), $(r_{\text{small}},r_{\text{large}})=(0.17,1)$, $n=3$, $p_{\max}=0.4$,
$\lambda=0.1$, and $k=0.5$.
We report both a single realization and an average over 50 independent runs.
The observed stability of the per-player objective trajectories is consistent with the theory:
the players’ actions evolve smoothly over time, reflecting bounded local regret and the
emergence of approximate first-order Nash equilibria.

\subsection{Mean--Variance--Skewness--Kurtosis Portfolio Design}

We next consider an online instance of the mean--variance--skewness--kurtosis (MVSK) portfolio
model \cite{PN11,BLP15,ZP21}, illustrating the applicability of \Cref{alg:4} to a nonconvex
online decision problem outside equilibrium-driven settings.

The objective is to allocate portfolio weights
$x\in\Delta_n:=\{x\in\mathbb{R}^n:\sum_{i=1}^n x_i=1,\;x_i\ge0\}$ by solving
\[
\min_{x\in\Delta_n}
\;-\mu_p(x)
+\frac{\lambda_2}{2}V_p(x)
-\frac{\lambda_3}{3}S_p(x)
+\frac{\lambda_4}{4}K_p(x)
+\frac{\lambda}{2}\|x\|_2^2,
\]
where $\mu_p$, $V_p$, $S_p$, and $K_p$ denote the portfolio mean, variance, skewness, and
kurtosis.

The problem is rendered online by estimating the required statistics from a rolling window of
past returns, inducing a time-varying and nonconvex objective.
The rolling-window estimation induces a setting in which repeated gradient evaluations
with respect to the same loss are natural and computationally inexpensive, making it well
suited for the nested oracle regime.
Using historical 100-day S\&P~500 price data, \Cref{alg:4} updates the portfolio at each time
step with a constant step size and a decreasing local-accuracy threshold
$\delta_t=(1+\sqrt{t})^{-1}$.

\begin{figure}[ht]
	\centering
	\begin{subfigure}[b]{0.45\textwidth}
		\centering
		\includegraphics[width=\textwidth]{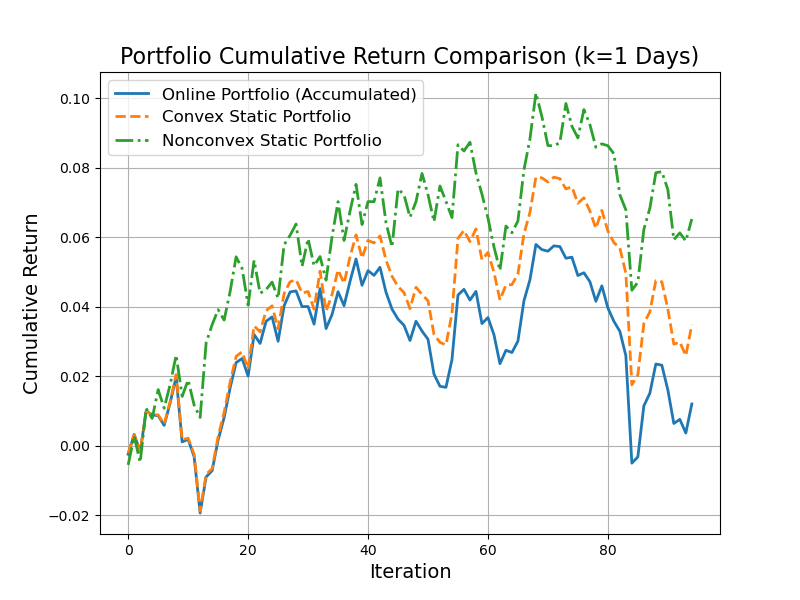}
		\caption{$k=1$.}
	\end{subfigure}
	\hfill
	\begin{subfigure}[b]{0.45\textwidth}
		\centering
		\includegraphics[width=\textwidth]{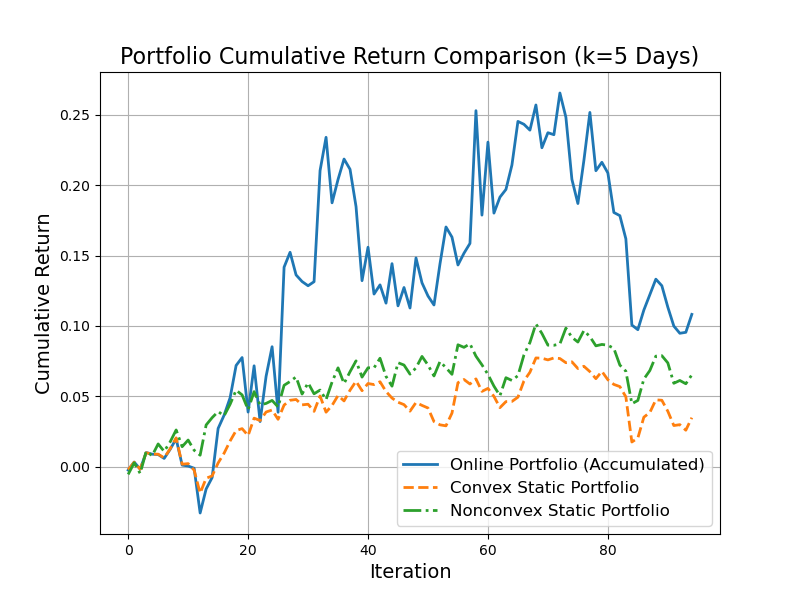}
		\caption{$k=5$.}
	\end{subfigure}
	\caption{Cumulative portfolio returns under the MVSK model.}
	\label{fig:portfolio}
\end{figure}

Performance is summarized via cumulative portfolio returns
\[
\bar{\br}_t
=
\prod_{i=1}^t\bigl(1+\langle r_i,x_i\rangle\bigr)-1.
\]
Results for window sizes $k=1$ and $k=5$ are reported in \Cref{fig:portfolio}.
For reference, we include two static benchmarks: a convex static portfolio obtained by
solving the mean–variance problem (by setting the higher-order MVSK parameters to zero),
and a nonconvex static MVSK portfolio, which serves as a proxy for the best static action
in hindsight.
These benchmarks act as offline comparators, isolating the effect of online adaptation
from static optimization quality.

Overall, the experiment illustrates that the proposed method adapts stably to a nonconvex,
time-varying objective, exhibiting consistent behavior despite relying exclusively on past
information and first-order updates.

\section{Conclusion}

This paper proposes a sound notion of local regret for nonconvex online  composite
optimization, grounded in first-order stationarity via proximal-gradient mappings.
In convex settings, this notion is equivalent, up to constants, to classical dynamic regret,
while in nonconvex settings it provides a meaningful performance measure that does not rely
on global optimality.
To the best of our knowledge, this is the first framework that establishes consistency and soundness of a regret criterion for nonconvex online  optimization.

The results highlight the central role of temporal regularity in online nonconvex
optimization and clarify the tradeoffs induced by different oracle-access models.
Beyond the general framework, the proposed notion admits natural interpretations in
structured settings such as nonconvex multiplayer games, where it corresponds to
approximate first-order Nash equilibria.

Several directions for future work arise naturally, including extensions to alternative
stationarity certificates (e.g., conditional-gradient, feasible-direction measures, Lagrangian-based conditions) and
specialized developments for application-driven models in operations research.
More broadly, the framework suggests that progress in nonconvex online  optimization hinges
on aligning regret notions with problem-specific stationarity concepts.

\section{Statements and Declarations}
\paragraph{Funding}
This work was supported by the ISRAEL SCIENCE FOUNDATION (grant No. 637/21).


\bibliographystyle{amsplain}
\bibliography{OPG_ref}
\end{document}